\documentclass[letterpaper, 10 pt, conference]{ieeeconf}

\IEEEoverridecommandlockouts                              

\usepackage{graphicx} 
\usepackage[cmyk]{xcolor}
\usepackage{epsfig} 
\usepackage{amsmath} 
\usepackage{amssymb}  
\usepackage{bm}
\usepackage{microtype}
\usepackage{placeins}
\usepackage{siunitx}
\usepackage{subcaption}
\usepackage{xcolor}
\usepackage{cite}
\usepackage{cancel}
\usepackage[symbol]{footmisc}
\usepackage{gensymb}
\usepackage{siunitx}
\usepackage{tabularx}
\renewcommand{\thefootnote}{\fnsymbol{footnote}}
\usepackage{flushend}
\usepackage{comment}
\usepackage{bbm}
\usepackage{balance}

\usepackage{multirow}
\usepackage{bigstrut}
\usepackage{booktabs}
\usepackage{pbox} 
\usepackage{makecell}
\usepackage{array}
\usepackage{tablefootnote}          
\usepackage{scrextend}              
\usepackage[normalem]{ulem}

\newcommand{\mytilde}{\raise.17ex\hbox{$\scriptstyle\mathtt{\sim}$}}
\newcommand{\myless}{\raise.27ex\hbox{\scriptsize\textless $\scriptstyle\mathtt{\sim}$}}
\newcommand{\mymore}{\raise.27ex\hbox{\scriptsize\textgreater $\scriptstyle\mathtt{\sim}$}}

\title{SpaceHopper: A Small-Scale Legged Robot for Exploring Low-Gravity Celestial Bodies}

\author{Alexander Spiridonov, Fabio Buehler, Moriz Berclaz, Valerio Schelbert, Jorit Geurts, Elena Krasnova, \\ Emma Steinke, Jonas Toma, Joschua Wuethrich, Recep Polat, Wim Zimmermann, Philip Arm, Nikita Rudin,\\Hendrik Kolvenbach, and Marco Hutter}

\begin{document}
\maketitle

\begin{abstract}
We present \emph{SpaceHopper}, a three-legged, small-scale robot designed for future mobile exploration of asteroids and moons. The robot weighs \SI{5.2}{\kg} and has a body size of \SI{245}{\mm} while using space-qualifiable components. Furthermore, \emph{SpaceHopper}'s design and controls make it well-adapted for investigating dynamic locomotion modes with extended flight-phases. Instead of gyroscopes or fly-wheels, the system uses its three legs to reorient the body during flight in preparation for landing. We control the leg motion for reorientation using Deep Reinforcement Learning policies. In a simulation of Ceres' gravity (\SI{0.029}{\g}), the robot can reliably jump to commanded positions up to \SI{6}{\m} away.   
Our real-world experiments show that \emph{SpaceHopper} can successfully reorient to a safe landing orientation within \SI{9.7}{\deg} inside a rotational gimbal and jump in a counterweight setup in Earth's gravity. Overall, we consider \emph{SpaceHopper} an important step towards controlled jumping locomotion in low-gravity environments. 
\end{abstract}
\begingroup
  \renewcommand\thefootnote{}\footnote{This research was supported in part by the Swiss National Science Foundation through the National Centre of Competence in Digital Fabrication (NCCR dfab). Corresponding author Alexander Spiridonov. Jonas Toma and Wim Zimmerman are with the ZHAW, 8401 Winterthur. All other authors are with the Robotic Systems Lab, ETH Zurich, 8092 Zurich, Switzerland. (e-mail: aspiridonov@ethz.ch; fabuehle@ethz.ch; mberclaz@ethz.ch; valerisc@ethz.ch; 
  jgeurts@ethz.ch;
  ekrasnova@ethz.ch; steinkee@ethz.ch; tomajon1@students.zhaw.ch; jwuethrich@ethz.ch; recep.polat@ethz.ch; zimmewim@students.zhaw.ch;  philip.arm@mavt.ethz.ch; rudinn@ethz.ch; hendrik.kolvenbach@mavt.ethz.ch; mahutter@ethz.ch;).
  © 2024 IEEE.  Personal use of this material is permitted.  Permission from IEEE must be obtained for all other uses, in any current or future media, including reprinting/republishing this material for advertising or promotional purposes, creating new collective works, for resale or redistribution to servers or lists, or reuse of any copyrighted component of this work in other works.}
  \addtocounter{footnote}{-1}%
  \endgroup

\begin{keywords}
Legged Robots, Space Robotics and Automation, Engineering for Robotic Systems 
\end{keywords}

\section{INTRODUCTION}
\label{sec_introduction}
Low-gravity celestial bodies like asteroids and moons are becoming increasingly popular targets for space missions. Their composition makes them valuable both from a scientific and economic perspective\cite{Elvis2012-uz}. Moreover, the dawn of a new Space Economy also promises to reduce the cost of visiting these objects. The launch of small-scale CubeSats has already become affordable for many enterprises\cite{new-space}. With this in mind, we aim to develop a robot enabling scalable mobile asteroid exploration. Traditionally used, wheeled locomotion systems, such as the Mars Rover Perseverance \cite{perseverence} or the proposed MMX Rover \cite{MMX}, are not very well suited for asteroids. Low gravity causes loss of wheel traction, and the rough terrain requires a very versatile system~\cite{kolvenbach2018scalability}. Other 
\begin{figure}[ht]
  \centering
  \includegraphics[width=0.97\columnwidth]{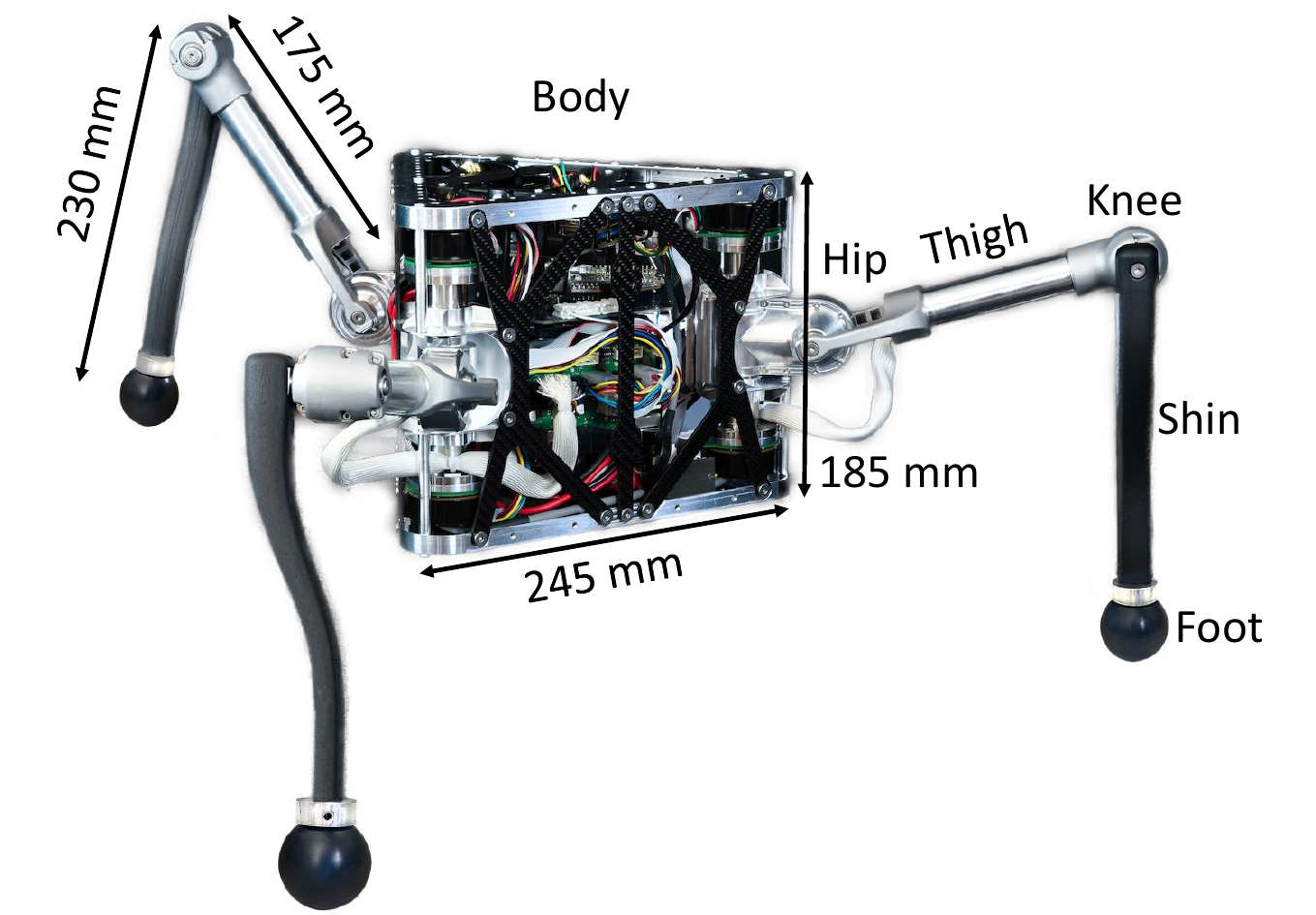}
  \caption{The small-scale low-gravity robot \emph{SpaceHopper} fully assembled with labeled parts and dimensions.}
  \label{fig:frontpage}
\end{figure}
robots, such as the asteroid hopper MASCOT \cite{ho2017mascot}, JAXAs Minerva \cite{Yabuta2019-xs} or legged climbing robots like LEMUR \cite{Lemur} and ReachBot \cite{ReachBot} offer alternative approaches to microgravity mobility. However, their non-articulated design or reliance on specialized grippers or docking adapters make them less suited for efficient, wide-ranging exploration of asteroids. Dynamically walking robots such as ANYmal\cite{anymal} and Spot\cite{autonomous_spot}, on the other hand, have shown impressive capabilities to traverse unstructured terrain on Earth. Thus, a variant of these robots has a high potential for mobile exploration of celestial bodies\cite{kolvenbach2021phd}. Some pathfinders, such as Spot Nebula \cite{nebula_spot} and SpaceBok\cite{Space-Bok}, have already been tested on analog sites but are not yet specifically engineered for space \cite{kolvenbach2021martianslopes}. Moreover, in the low-gravity environment of asteroids, gaits with extended flight phases are increasingly efficient \cite{kolvenbach2018efficient}. Dynamic jumping locomotion is also advantageous
compared to other walking gaits, as it can swiftly overcome obstacles by simply jumping over them. However, it also poses the challenge of stabilizing the robot's attitude in low-gravity during the jump phases \cite{kolvenbach2019iros}. Traditionally, attitude control in space and on low-gravity celestial bodies uses reaction wheels. However, to save weight and eliminate the complexity of an additional subsystem, one can omit reaction wheels and control the robot's attitude using the legs only. SpaceBok, which only has two degrees of freedom (DOF) per leg, used this idea to perform jumping locomotion in a planar representation of a microgravity environment \cite{nrudin_catlikejumping}.  

In this paper, we present \emph{SpaceHopper}, a small-scale, lightweight, and space-qualifiable robot to investigate controlled locomotion in low-gravity scenarios. The proposed system uses jumping as the primary form of locomotion. Moreover, the robot can control its attitude in low-gravity using solely its legs. Besides demonstrating the mechanical and electrical feasibility of the system, we demonstrate SpaceHopper's jumping and reorientation capability in simulation and partly on the real system. Specifically, we provide the following contributions:
\begin{itemize}
    \item We present a three-legged, CubeSat-sized, and lightweight robot specifically designed for controlled low-gravity locomotion. 
    \item In a simulation, we validate the \emph{SpaceHopper} design by showing accurate low-gravity attitude control and jumping to commanded positions.
    \item We demonstrate 2D attitude control and vertical jumping on the hardware using a gimbal and a counterweight setup.
    
\end{itemize}

\section{SYSTEM DESIGN}
\label{sec_system_design}
\subsection{System Overview}
\emph{SpaceHopper} is a three-legged robot used as a research platform for low-gravity locomotion in space. We tested \emph{SpaceHopper} in a lab environment (see Sec. \ref{sec:attitude_control} and \ref{sec:pronking_locomotion}) and accordingly estimate it to be at technology readiness level (TRL) four. 
The body is triangular and symmetric, with one leg attached to each corner (Fig.~\ref{fig:frontpage}).
\emph{SpaceHopper} can fit into a 27U CubeSat \cite{CubeSats} with its legs fully tucked in. At \SI{5.2}{\kg}, the robot is significantly lighter than the \SI{54}{\kg} CubeSat limit.

\begin{figure}[h]
    \centering
    \includegraphics[width=0.9\columnwidth]{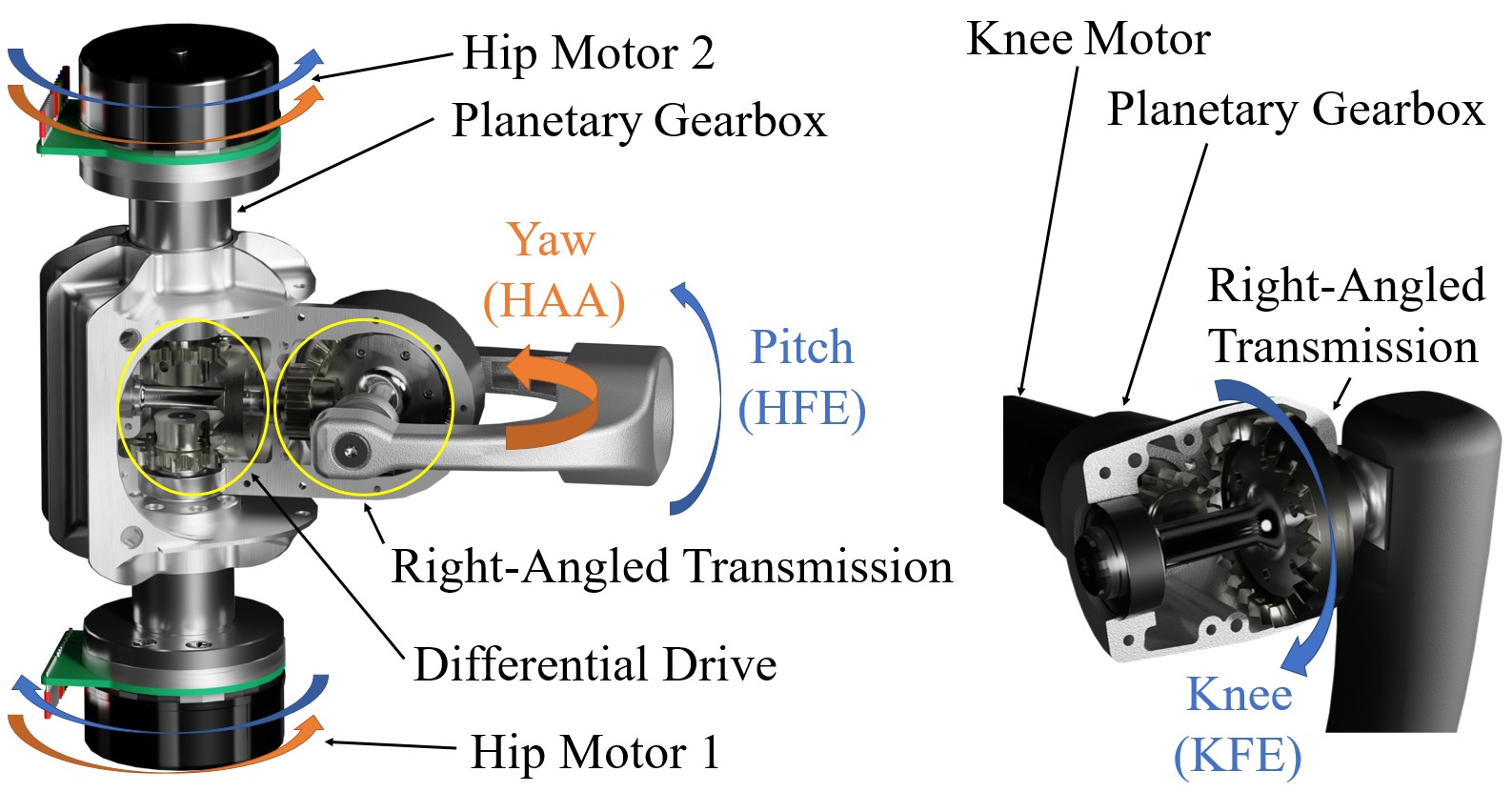}
    \caption{Cut section of the hip and knee with annotated transmissions, movements, and components.}
    \label{fig:Whole_Leg_annotated}
\end{figure}
\subsection{Leg Design}
Jumping and reorientation is the main locomotion goal of \emph{SpaceHopper} and a guiding feature for the leg design.
Having three legs is a unique feature of \emph{SpaceHopper} and allows for weight saving compared to a classic four-legged design (one leg weighs \SI{1.17}{\kg}, including all motors). Static standing is possible with three legs, which increases stability compared to a two-legged design. Four legs enable static locomotion, but given the low-gravity environment, this gait is less efficient \cite{kolvenbach2018efficient} and not favorable.
\emph{SpaceHopper's} legs each have three active DOFs as this vastly outperforms legs with only two active DOFs in attitude control \cite{nrudin_catlikejumping}. Two of those active DOFs are in the hip and are realized through a differential drive (Sec.~\ref{subsec_hip_design}), which enables hip flexion/extension (HFE) and hip abduction/adduction (HAA). The other active DOF is placed in the knee, enabling flexion/extension (KFE).
Here, we used a classical right-angled drive to place the knee motor efficiently in the thigh (see Fig. \ref{fig:Whole_Leg_annotated}).
The chosen actuators are Maxon EC 45 flat \cite{EC45flat} in the hip and ECX Torque 22 M \cite{ECXTorque22M} in the knee. We chose these motors for their high torque-to-weight/size ratio and the availability of space-graded versions with the same specifications. The resulting maximum joint torques, accounting for the transmission ratios, are shown in Table \ref{tab:max_torques}.
As depicted in Fig. \ref{fig:Whole_Leg_annotated}, we chose a pinion-crown gear combination instead of classic bevel gears for all right-angled transmissions because pinion-crown gear combinations allow for looser tolerances in axial positioning \cite{kissling2007face}. This freedom is advantageous, considering the large temperature fluctuations in space. 
\begin{table}[h]
    \centering
    \renewcommand{\arraystretch}{1.5}
    \setlength{\tabcolsep}{.118cm} 
    \begin{tabular}{l|cc|cccc|}
        \cline{2-7}
        & \multicolumn{2}{c|}{\textbf{Knee}} & \multicolumn{4}{c|}{\textbf{Hip}} \\
        \hline
        \multicolumn{1}{|l|}{} & \textbf{Motor} & \textbf{Joint} & \textbf{Motor} & \textbf{Pitch \& Yaw} & \textbf{Pitch} & \textbf{Yaw} \\
        \hline
        \multicolumn{1}{|l|}{\textbf{Max torque \SI{}{\N\m}}} & 0.078 & 0.740 & 0.152 & 1 & 2.8 & 1.5 \\
        \hline
    \end{tabular}
    \caption{Maximum achievable torques with chosen motors, as limited by winding temperature, power, and stall torque.}
    \label{tab:max_torques}
\end{table}
\subsection{Hip Design}
\label{subsec_hip_design}
Many modern quadrupeds like ANYmal \cite{anymal} by ANYbotics or Boston Dynamics' Spot \cite{spot} use a serially linked drivetrain. This drivetrain has a simple mechanical design and high efficiency \cite{abate2016mechanical}. Nevertheless, for \emph{SpaceHopper's} special use case in low-gravity environments, we chose a parallel linked drivetrain in the form of a differential drive for the hip actuation based on the following considerations:
\subsubsection{Combined Torque for Jumping}
The differential drive allows both motors to contribute to the pitch torque in all configurations, in contrast to a serially linked drivetrain. \emph{SpaceHopper} can benefit significantly from the differential drive, as jumps rely on the HFE motion for which the motors can work in conjunction and provide double the torque (see Fig. \ref{fig:Whole_Leg_annotated}).
For combined pitch and yaw motions, the torques are split up according to Equation \ref{eqn:Hip}, which is not critical, as combined motion only occurs during reorientation when the loads are not significant.
\begin{equation}
    \begin{pmatrix} \gamma_{h1} \\ \gamma_{h2} \end{pmatrix} =  \begin{pmatrix} 1 / (2 i_{hd}i_{hr} i_{hd}) & 1/ (2 i_{hp})\\  -1 / (2  i_{hd}  i_{hr}  i_{hp})& 1 / (2  i_{hp}) \end{pmatrix} \cdot \begin{pmatrix} \tau_{p} \\ \tau_{y} \end{pmatrix}
    \label{eqn:Hip}
\end{equation}
\noindent The torques of the two hip motors and the pitch, yaw joint are $\gamma_{h1}, \gamma_{h2}, \tau_{p}, \tau_{y}$. The gear ratios of the differential, the right-angled, and the planetary transmission are indicated by $i_{hd}, i_{hr}, i_{hp}$, respectively.

\subsubsection{Static Motor Placement}
Using a differential drive in the hip allows us to place two motors per leg inside the body. This placement of the motors is very space-efficient and eases thermal management, dust protection, and cable management for space applications.

In order to achieve the same range of motion (ROM) in hip flexion/extension as offered by a serially linked drivetrain, the differential drive is paired with a right-angled, second-stage transmission, as shown in Fig.~\ref{fig:Whole_Leg_annotated}. 
The resulting ROM is $\pm \SI{85}{\degree}$, which allows \emph{SpaceHopper} to develop higher take-off velocities while jumping off. The hip abduction/adduction has a ROM of $\pm \SI{70}{\degree}$, while the knee has $\pm \SI{170}{\degree}$.
The second stage transmission also enables fine-tuning the torque of the respective hip motions independently. As jumping requires higher torque than reorientation, \emph{SpaceHopper's} hip pitch motion has twice the torque of the hip yaw motion ($i_{hr} = 2$).

\subsection{Body Design}
The triangular body shape, maximizes the ROM in the yaw DOF while minimizing the size of the body to save weight.
The internal structure is 3D printed, which allows for high design flexibility. Thanks to this flexibility, we can tightly integrate the body internals and use vibration dampers to decouple the electronics stack from the load-bearing structure. 

\subsection{Materials}
Most of the structural parts in the legs and the main triangles of the body are manufactured out of aerospace aluminum 7075, which has a good strength-to-weight ratio and is commonly used in space applications \cite{NASA_Space_Aluminum}.
The shin and the body's internal electronics structure consists of 3D printable carbon-reinforced composites, which strikes a good tradeoff between the cost of manufacturing and structural robustness.
The gears and shafts are manufactured from hardened steel (16MnCr5 and 30CrNiMo8, respectively) to endure the high surface pressures and bending forces.
Finally, the side structure is water-jet cut out of \SI{2}{\mm} thick, layered carbon fiber for high tensile strength and low weight.

\subsection{Electrical System}
\label{subsec:electrical_system}
In the center of the electronics stack is a 7S1P lithium-ion battery pack managed by a Tiny BMS s516 v2.1 battery management system~\cite{Enepaq}. We use Lithium-ion cells because of their high power-to-weight ratio.
A custom-built power distribution board (PDB) distributes the power from the battery to the motors and the onboard processing unit (Nvidia Jetson Nano).
An STM32 microcontroller \cite{STM32} connects to an internal temperature sensor, the BMS, and the Nvidia Jetson Nano. The STM32 chip shuts down the system in case of overheating or other failures, which adds to the system's robustness.
We use the Nvidia Jetson Nano because it allows for efficient inference of the control policy via its GPU cores.
We use three Maxon EPOS4 Compact 24/5 EtherCAT 3-axes motor controllers \cite{MotorController}.
They are compact and support regenerative braking, which increases efficiency and prolongs mission duration.
The robot also has three time-of-flight laser range sensors (LRFs) \cite{LRF}, each with a maximum range of \SI{4}{\m}, delivering height information.

\vspace{8pt}
\section{CONTROL SYSTEM OVERVIEW}
\label{sec_controls}

\subsection{Control Architecture}
\label{subsec:control_architecture}
\begin{figure}[h]
    \centering
    \includegraphics[width=0.9\columnwidth]{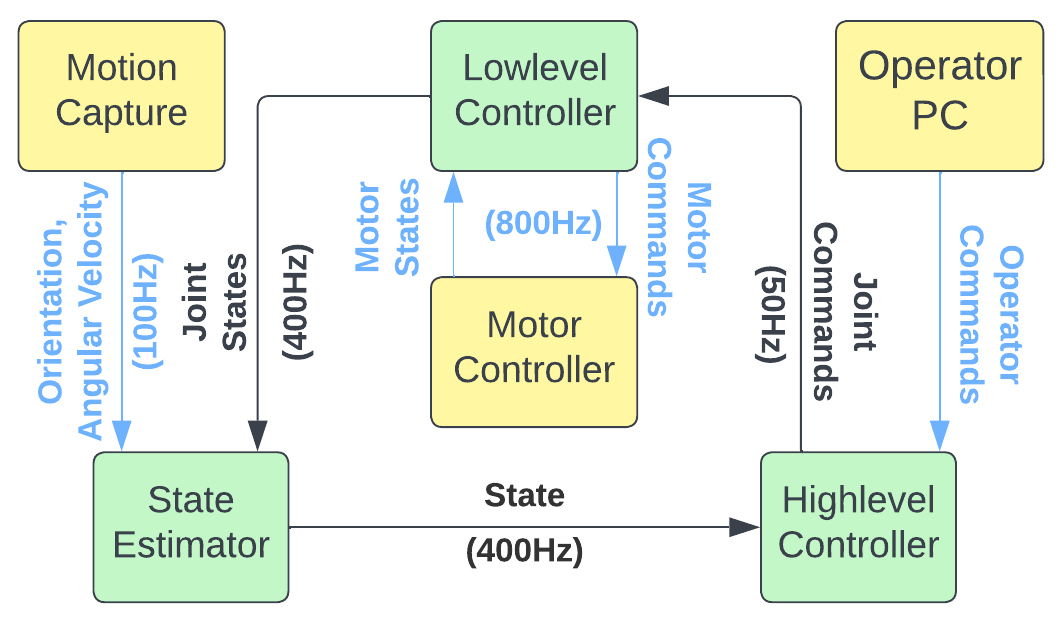}
    \caption{Dataflow visualization inside \emph{SpaceHopper} during reorientation test. Green boxes show software modules running on the onboard computer, and yellow boxes show peripheral modules.}
    \label{fig:control_architecture}
\end{figure}

The control framework of \emph{SpaceHopper} uses shared memory buffers that communicate with the peripherals via the Robot Operating System (ROS) \cite{ros}. The control framework comprises three parts: a low-level controller, a high-level controller, and a state estimator.
The state estimator receives joint states, motion capture data, and laser-ranging sensor information and uses it to estimate the height and attitude of the main body (see Fig. \ref{fig:control_architecture}).
The high-level controller processes the data received by the state estimator and computes desired joint positions. These desired joint positions are sent to the low-level controller, which maps them to desired motor positions and sends them to the motor controllers through an EtherCAT network.

\subsection{Locomotion Controller}
Developing a robust controller for low-gravity jumping locomotion is challenging. Due to the extended flight phases, small torques on the body during jump-off compound to large changes in orientation. For a safe landing, controlling the robot's attitude during the flight phase is paramount. However, reorientation using the legs is difficult. The kinematics of a free-floating space manipulator are nonholonomic, and dynamic singularities can significantly restrict the controllable workspace \cite{Papadopoulos1993}. Model-based control methods exist that compute approximately optimal trajectories using non-linear optimization. However, they are computationally slow and do not scale well to multiple legs \cite{Fernandes}. Other methods avoid the dynamic singularities by applying small circular motions with the leg \cite{Vafa1993}. However, these solutions result in prolonged reorientation times. On the other hand, model-free Deep Reinforcement Learning (DRL) has already been used successfully for free-floating attitude control of the robot SpaceBok \cite{nrudin_catlikejumping}.
Thus, we build upon this previous result and use a modified version of the proximal policy optimization (PPO) algorithm \cite{schulman_ppo, nrudin_learningtowalk} to train locomotion policies for \emph{SpaceHopper}. 
The resulting locomotion policies are neural networks that take observations as inputs and output desired joint positions. The policy's architecture is a Multi-Layer Perceptron (MLP) with three hidden layers of size $[512, 256, 128]$. 

Typically, DRL-based locomotion controllers are trained in simulation and then transferred to the real world using sim-to-real techniques \cite{simtoreal_survey}. We train the policies in the IsaacGym simulation environment \cite{isaacgym} for fast training, thanks to its GPU acceleration. To transfer the policies to the real robot, we employ domain randomization \cite{domain_rand} of the robot's link masses, the center of mass, DOF frictions, and P\&D-Gains.

\begin{center}
    \begin{table}[h]
        \begin{tabularx}{\columnwidth}{l l}
            \hline\\
            Policy & Observations\\[0.2cm]
            \hline\\
            Attitude Control Simulation & $\mathbf{q}$, $\mathbf{\Dot{q}}$, $\mathbf{q}_b$, $\boldsymbol{\omega}_b$, $\mathbf{a}_{t-1}$\\[0.2cm]
            Attitude Control Gimbal & $\mathbf{q}$,  $\mathbf{q}_b$, $\boldsymbol{\omega}_b$, $\mathbf{a}_{t-1}$\\[0.2cm]
            Jumping Simulation & $\mathbf{q}$, $\mathbf{\Dot{q}}$, $\mathbf{r}_b$, $\mathbf{r}^*_b$, $\mathbf{v}_b$, $\mathbf{q}_b$, $\boldsymbol{\omega}_b$, $\mathbf{a}_{t-1}$\\[0.2cm]
        \end{tabularx}
        \begin{tabularx}{\columnwidth}{r l r l}
            \hline\\
            orientation\_3d: & $\|\textit{rot\_vec($\mathbf{q_b}$)}\|_2$ & orientation\_2d: & $7-(|\theta| + |\varphi|)$\\[0.2cm]
            action\_rate: & $\|\mathbf{a}_t - \mathbf{a}_{t-1}\|_2$ & torques: & $\|\boldsymbol{\tau}\|_2^2$\\[0.2cm]
            dof\_vel: & $\|\mathbf{\Dot{q}}\|_2$ & dof\_acc: & $\|\mathbf{\Ddot{q}}\|_2$\\[0.2cm]
        \end{tabularx}
        \begin{tabularx}{\columnwidth}{r l}
        $\quad$ \,   collision: & $\sum_j \mathbbm{1}\left( \| \mathbf{f}_j \|_2 > 0.2 \right)$ $\quad$ \hspace{2pt} height: $\quad$ $|r_{b,3}|$\\[0.2cm]
        $\quad$ \,   pos\_cmd: & $1-\|\mathbf{r}_b - \mathbf{r}_b^*\|_2 / \|\mathbf{r}_b^*\|_2$\\[0.2cm]
        $\quad$ \,   dof\_limits: & $\| \text{ReLU}( \mathbf{q}_l - \mathbf{q} ) \|_1 + \| \text{ReLU}( \mathbf{q} - \mathbf{q}_u ) \|_1 $ \\[0.2cm]
         \end{tabularx}
         \begin{tabularx}{\columnwidth}{r l r l}
            \toprule
            $\mathbf{q}$: & Joint Positions & $\mathbf{\Dot{q}},  \mathbf{\Ddot{q}}$: & Joint Velocities and Accelerations\\[0.2cm]
            $\mathbf{r}_b$: & Body Position & $\mathbf{v}_b$: & Body Velocity\\[0.2cm]
            $\mathbf{q}_b$: & Body Quaternions & $\boldsymbol{\omega}_b$: & Body Angular Velocity\\[0.2cm]
            $\mathbf{a}_{t}$: & Actions& $\mathbf{r}^*_b$: & Commanded Body Position\\[0.2cm]
            $\mathbf{q}_l$: & Lower DOF Limit & $\mathbf{q}_u$: & Upper DOF Limit\\[0.2cm]
            $\boldsymbol{\tau}$: & Joint Torques& $\mathbf{f}_j$: & Contact Forces of Body $j$\\[0.2cm]
            \hline
        \end{tabularx}
        
        \caption{The policy observations and reward components.}
        \label{tab:rewards_observations}
    \end{table}
\end{center}
\vspace{-29pt}
\section{ATTITUDE CONTROL}
\label{sec:attitude_control}
Since low gravity attitude control is a critical capability, we investigate it separately. In a first step, we train a policy using DRL to reorient while freely floating in zero gravity. We then validate the reorientation capability on the real system. Creating an accurate low gravity reorientation test in Earth's gravity for a system with 3DOF per leg is challenging. To this end, we developed a custom gimbal test rig that offloads the gravity from the body while allowing it to rotate freely. We also train a DRL policy to reorient in this test setup. While this gravity offload system substantially alters the robot's dynamics, it allows us to test the mechanical, electrical and control systems during attitude control maneuvers. Moreover, it helps us assess the sim to real performance.

\subsection{Simulation Study}

\subsubsection{Experiment Setup}
Since the robot effectively experiences weightlessness between jump-off and landing, we simulate the robot floating in zero gravity. The robot starts in a random initial orientation and then reorients to an upright landing configuration.

\subsubsection{Controller}
We use a DRL policy.
All policy observations and reward components are listed in Table \ref{tab:rewards_observations}.
The total reward used during training is: 
\begin{equation}
\begin{split}
    r = c_1 \cdot r_{\text{orientation\_3d}} + c_2 \cdot r_{\text{action\_rate}}\\
    + c_3 \cdot r_{\text{torques}} + c_4 \cdot r_{\text{dof\_limits}}
    \end{split}
\end{equation}
Where $(c_1, c_2, c_3, c_4) = (-1,-0.04,-0.15, -3)$.
Here, $r_{\text{orientation\_3d}}$ penalizes non upright attitudes. $r_{\text{action\_rate}}$ prevents high-frequency oscillations in the joints, $r_{\text{torques}}$ encourages the policy to find a low-torque solution and $r_{\text{dof\_limits}}$ penalizes joint positions close to the ROM limit.

\subsubsection{Results}
\begin{figure}[h]
    \centering
    \includegraphics[width=0.7\columnwidth]{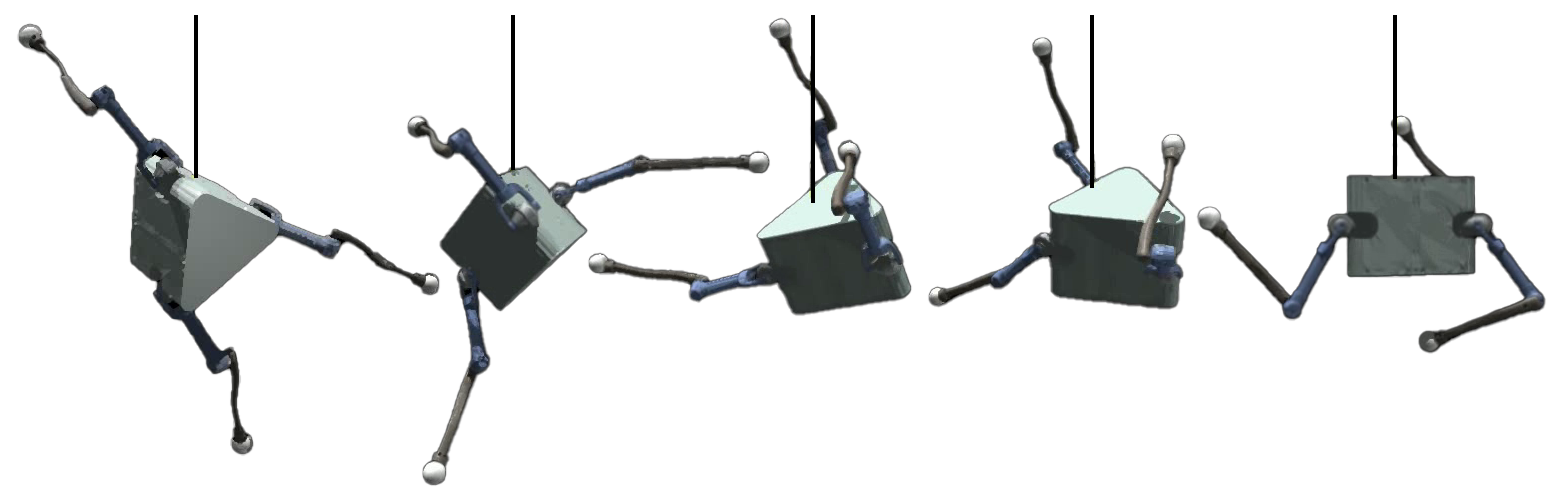}
    \caption{\emph{SpaceHopper} reorienting itself to the upright orientation, which is displayed as a black line.}
    \label{fig:reorientation_illustration}
\end{figure}

\begin{figure}[h]
  \centering  \includegraphics[width=1\columnwidth]{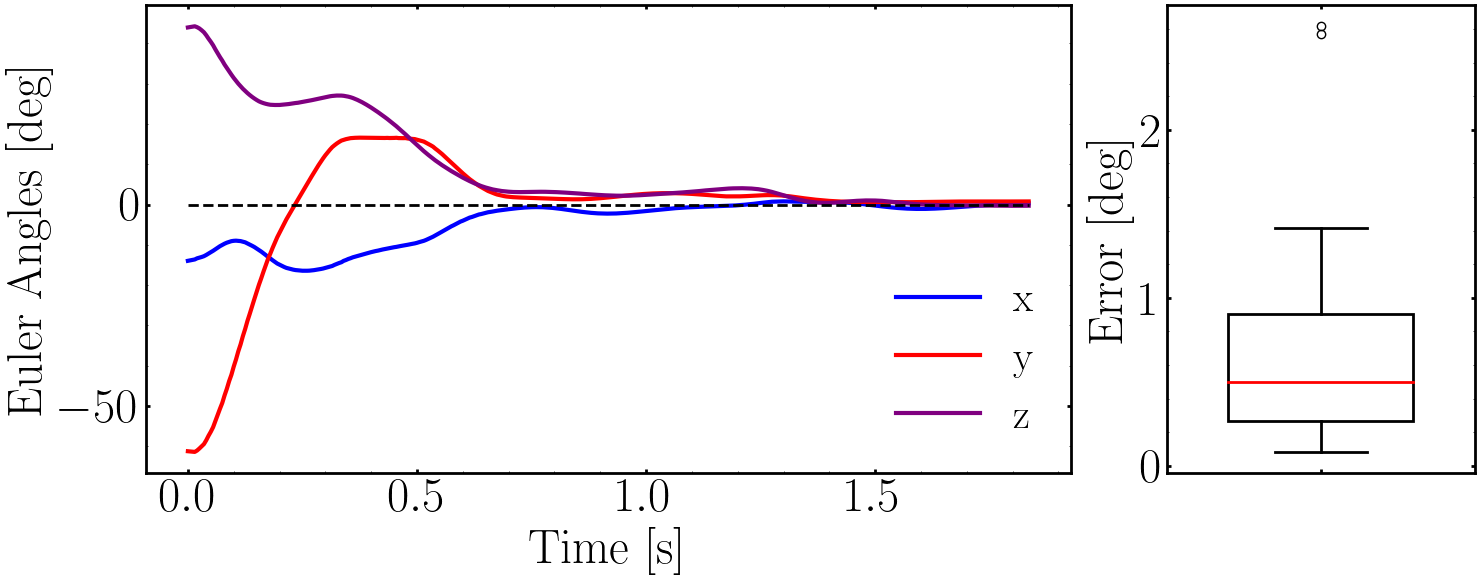}
  \caption{\emph{SpaceHopper's} body orientation (Euler angles $x,y,z$) in simulation during reorientation from random initial attitude to upright (left panel), and a box plot of final orientation errors for twenty random initial orientations (right panel).}
  \label{fig:results_reorientation_3d}
\end{figure}

After training, the robot can reorient itself to an upright orientation precisely and rapidly. Fig. \ref{fig:reorientation_illustration} shows the robot changing its attitude from a random initial orientation to upright. The policy reaches an upright orientation within \SI{1}{\s} (see left panel of Fig. \ref{fig:results_reorientation_3d}). Moreover, the attitude control is robust, with a mean smallest angle orientation error of $\SI{0.501}{\deg}$ for twenty random initial orientations, as seen in the right panel of Fig. \ref{fig:results_reorientation_3d}. Overall, considering that a jump with a distance of \SI{6}{\m} in Ceres gravity (\SI{0.029}{\g}) takes around \SI{8}{\s} (see Fig. \ref{fig:results_torques}), the 
reorientation is fast enough to reach an upright landing configuration.

\vspace{-18pt}

\subsection{Testing on Hardware}
\label{subsec:testing_on_hardware}
\subsubsection{Experiment Setup}
\begin{figure}[h]
    \centering
    \includegraphics[width=0.6\columnwidth]{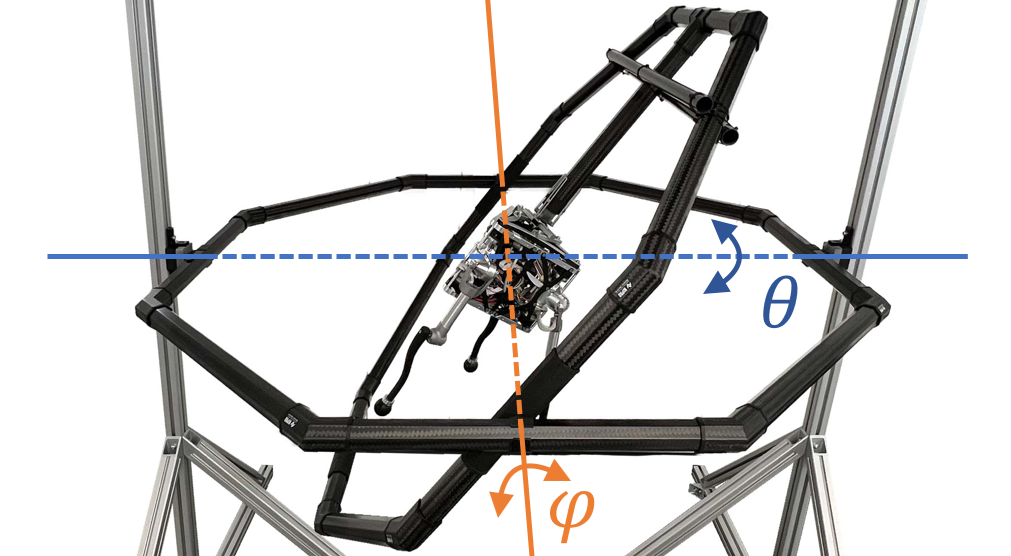}
    \caption{SpaceHopper mounted in the gimbal test stand. The inner ring angle is $\varphi$ in orange, and the outer ring angle $\theta$ in blue.}
    \label{fig:gimbal_setup}
\end{figure}

We test the capabilities of \emph{SpaceHopper} 
using a custom gimbal, allowing the robot to rotate freely in two degrees of freedom. The inner ring can rotate with the angle $\varphi$ and the outer ring with angle $\theta$ (see Fig. \ref{fig:gimbal_setup}). The robot starts in a random initial configuration of $\varphi$, $\theta$ and has to reach an upright attitude. We track the body orientation and angular velocity using a Vicon motion capture system.  

\subsubsection{Controller}
We use a DRL policy.
The total reward is:
\begin{equation}
\begin{split}
    r = c_1 \cdot r_{\text{orientation\_2d}} + c_2 \cdot r_{\text{action\_rate}} + c_3 \cdot r_{\text{torques}}&\\
    + c_4 \cdot r_{\text{dof\_limits}} + c_5 \cdot r_{\text{dof\_acc}} + c_6 \cdot r_{\text{dof\_vel}}&
\end{split} 
\end{equation}
Where $(c_1, c_2, c_3, c_4, c_5, c_6)=(0.15,-0.06,-0.01,-1,-4\cdot10^{-6}, -0.01)$.
$r_{\text{orientation\_2d}}$ encourages the robot to be upright. We use $r_{\text{dof\_vel}}$ and 
$r_{\text{dof\_acc}}$ to avoid sim-to-real problems.
The environments terminate when the robot collides with itself or with a part of the gimbal.

\subsubsection{Results}
\begin{figure}[h]
    \centering
    \includegraphics[width=\columnwidth]{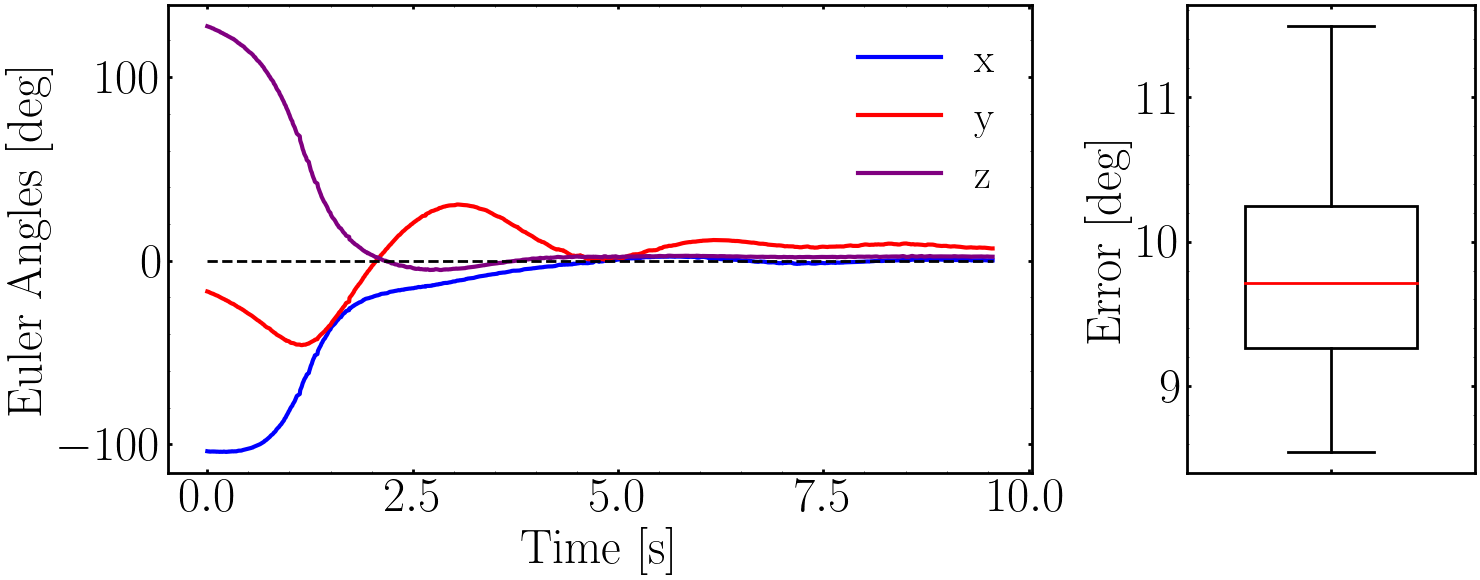}
    \caption{\emph{SpaceHopper's} body orientation (Euler angles $x,y,z$) in the gimbal during reorientation from random initial attitude to upright (left panel), and a box plot of final orientation errors for twenty random initial orientations (right panel).}
    \label{fig:reorientation_gimbal_panels}
\end{figure}

We analyze the performance of \emph{SpaceHopper's} reorientation capability by releasing the robot at twenty random initial configurations of the gimbal angles $\varphi$ and $\theta$. \emph{SpaceHopper} reaches an upright attitude from all configurations. The left panel of Fig. \ref{fig:reorientation_gimbal_panels} shows the change of attitude for one random initial configuration. It takes the robot \SI{5}{\s} to reach the upright orientation. The mean final orientation error is \SI{9.7}{\deg} over twenty random initial configurations (see right panel of Fig. \ref{fig:reorientation_gimbal_panels}). Despite the significant dynamics mismatches, which we further elaborate on in the Discussion \ref{subsec:limitations_of_test_scenarios}, the promising results of this experiment serve as a positive indicator for \emph{SpaceHopper's} capabilities in future more realistic micro-gravity experiments.

\vspace{-20pt}

\section{JUMPING LOCOMOTION}
\label{sec:pronking_locomotion}
Demonstrating controlled low-gravity jumping locomotion is the overarching goal of \emph{SpaceHopper}. Integration of jumping, attitude control, and landing is necessary for successful jumping locomotion. We first demonstrate controlled jumping locomotion in a simulation of Ceres (\SI{0.029}{\g}). Since validation of this controller on the real hardware in Earth's gravity requires extensive testing facilities \cite{bekdash2020development}, we restrict ourselves to showing the real robot's jumping capability only for vertical jumping in a counterweight setup.

\subsection{Simulation Study}
\label{subsec:simulation_study}

\subsubsection{Experiment Setup}

The simulation environment consists of a flat ground plane with the gravity of Ceres. The robot starts on the ground and is commanded to jump to a goal body position $\mathbf{r}_b^*$ with radius \SI{6}{\m} around the initial position. 

\subsubsection{Controller}
We propose an end-to-end approach for jumping, reorientation, and landing. A DRL policy trained simultaneously for all three tasks is not limited by heuristic transitions between controllers. The total reward is: 
\begin{equation}
\begin{split}
    r = c_1 \cdot r_{\text{pos\_cmd}} + c_2 \cdot  r_{\text{orientation\_3d}}  + c_3 \cdot r_{\text{action\_rate}}&\\  + c_4 \cdot r_{\text{torques}} + c_5 \cdot r_{\text{collision}} +c_6 \cdot r_{\text{height}}&
\end{split}
\end{equation}

The reward component weights have the following values: $(c_1, c_2, c_3, c_4, c_5, c_6)=(1.5,-0.4,-0.05,-0.4, -15, 0.19)$.
Here, $r_{\text{pos\_cmd}}$ encourages reducing the distance to the goal position, $r_{\text{collision}}$ penalizes collisions with the ground, $r_{\text{orientation\_3d}}$ enforces an upright body attitude and $r_{\text{height}}$ incentivizes the exploration of solutions involving jumping.
 
\subsubsection{Results}
\begin{figure}[h]
  \centering
  \includegraphics[width=1\columnwidth]{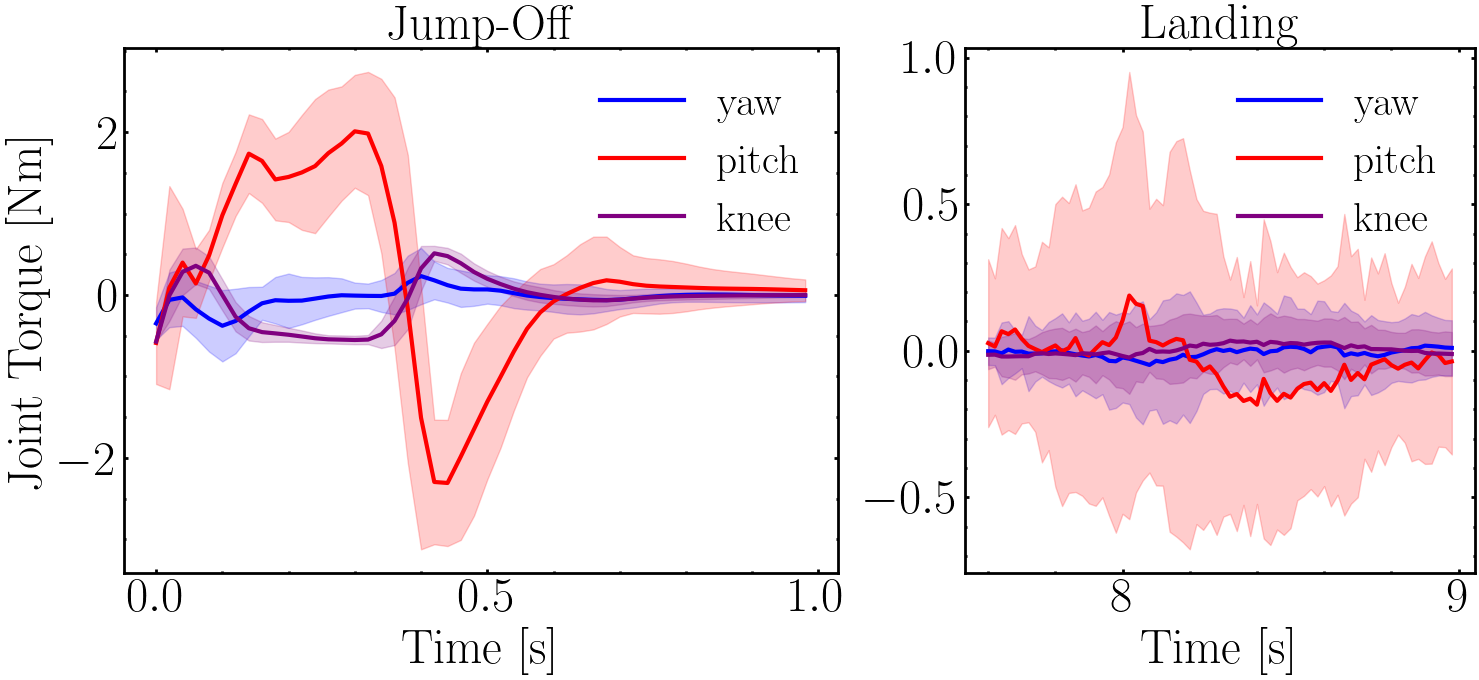}
  \caption{The yaw, pitch, and knee joint torques during jumps to 100 random positions in simulation. The thick lines are the mean and the shaded areas are the standard deviations. We average the joint torques from different legs. The robots leave the ground around the \SI{0.4}{\s} mark.}
  \label{fig:results_torques}
\end{figure}

\begin{figure}[h]
  \centering
  \includegraphics[width=0.93\columnwidth]{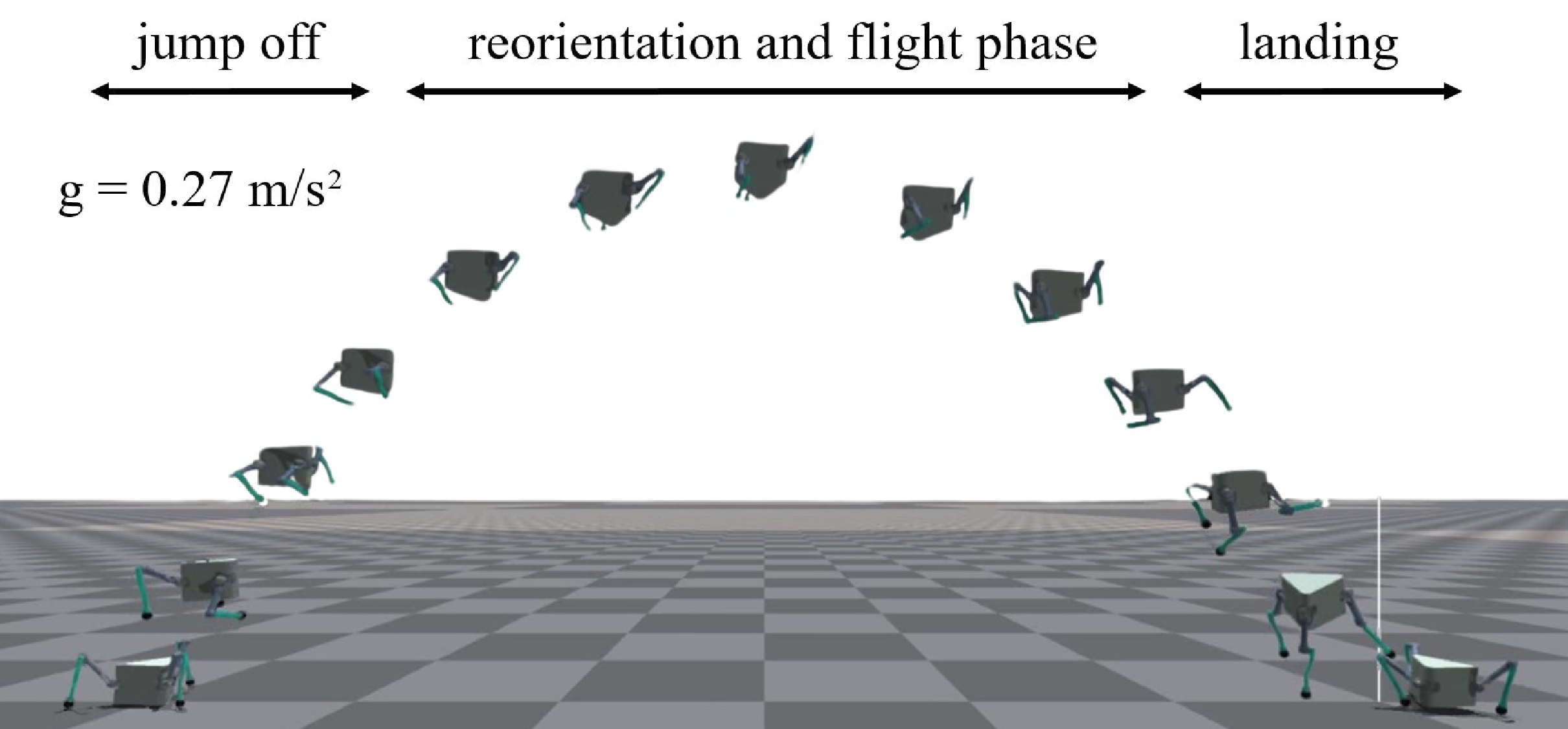}
  \caption{\emph{SpaceHopper} in Ceres gravity, jumping to a \SI{6}{\m} far commanded position (white line).}
  \label{fig:results_jumping_ceres}
\end{figure}

As shown in Fig. \ref{fig:results_jumping_ceres}, this policy can handle jumping, reorientation, and landing end-to-end. The robot reaches a maximum height of \SI{2.55}{\m}, stabilizes its attitude, and manages to land on its feet within \SI{9}{\s}. For 100 random position commands at a distance of \SI{6}{\m}, the policy achieves an average position error of \SI{0.316}{\m}. The worst case position error is \SI{0.843}{\m}. Most importantly, the robot's body, thighs, and shins never collide with the ground. Figure \ref{fig:results_torques} confirms our assumption that jumping requires higher pitch than yaw torques.

\vspace{-15pt}
\vspace{-10pt}
\subsection{Testing on Hardware}
\label{subsec:pronking_locomotion:testing_on_hardware}
\subsubsection{Experiment Setup}
\emph{SpaceHopper's} drivetrain is not powerful enough to allow jumping on Earth. We use a simple pulley-counterweight setup that attaches to the top of the robot's main body to offload the gravity from the body.
Using a counterweight of \SI{3.9}{\kg}, we simulate a gravity of \SI{2.5}{\m/\s^2} on the body's vertical axis. We chose this value so that the robot does not exceed the maximum height of the test stand while jumping.

\subsubsection{Controller}
Due to the limitations of the test setup (see Subsec. \ref{subsec:limitations_of_test_scenarios}), we only validate the jumping capability of the mechanical and electrical systems. Accordingly, we omit a full DRL-based control approach and use a simple controller that tracks a hand-crafted joint-space trajectory with the height estimation from the laser-ranging sensors.
The trajectory consists of a leg extension to initiate the jump and a contraction during the flight phase to prepare for the next jump.
This trajectory is the same for all three legs and gets triggered when the robot is below a certain height.

\subsubsection{Results}

\begin{figure}[h]
  \centering   \includegraphics[width=0.9\columnwidth]{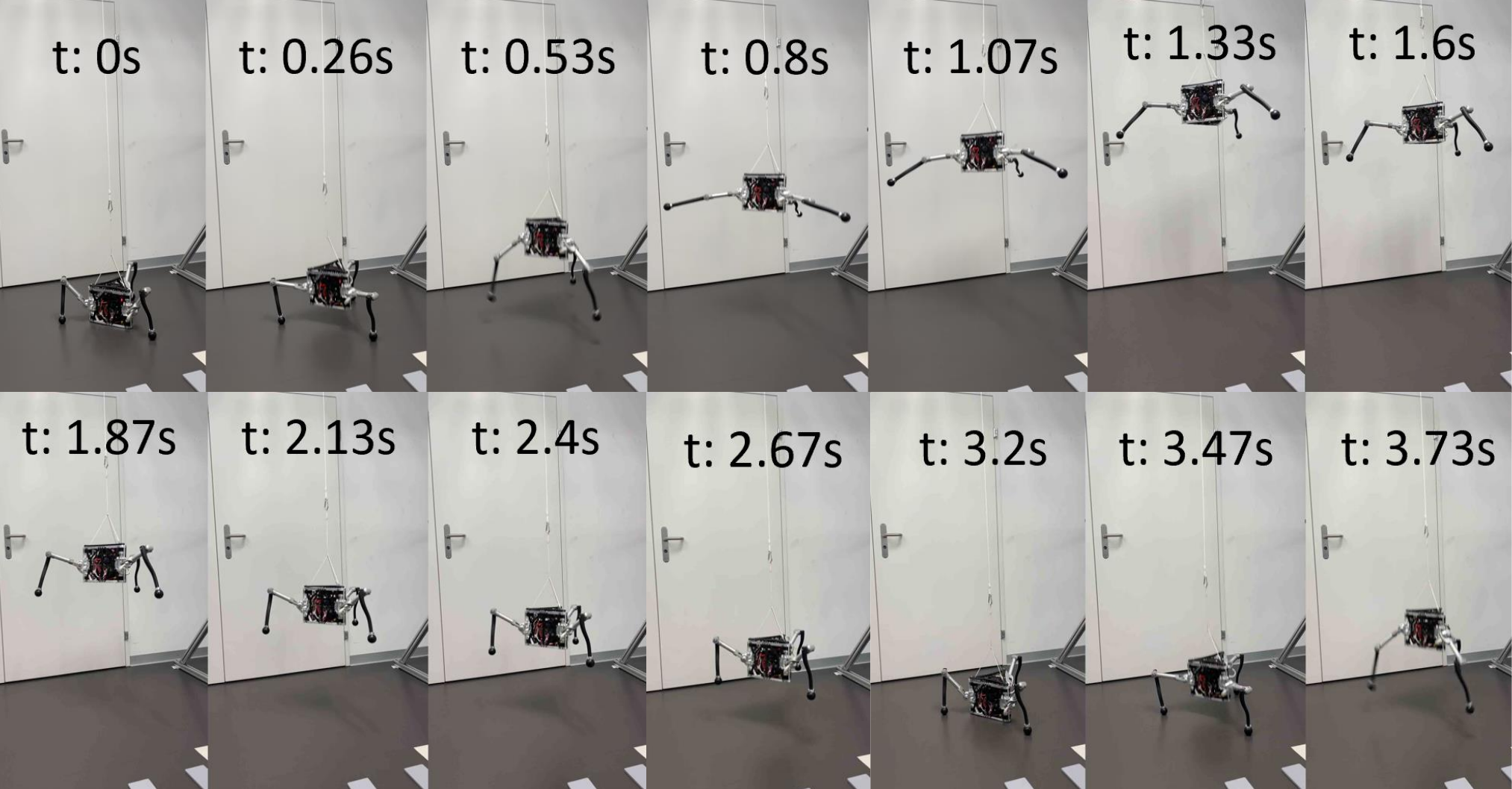}
  \caption{\emph{SpaceHopper} jumping continuously with a counterweight of \SI{3.9}{\kg} being attached to the top of the body over a pulley system.}
  \label{fig:results_jumping}
\end{figure}

Fig. \ref{fig:results_jumping} shows \emph{SpaceHopper} jumping to a height of \SI{1.2}{\m}, landing and immediately jumping off again. In this continuous jumping mode, \emph{SpaceHopper} achieves at least four and up to 15 consecutive jumps before a human has to intervene. The failure mode is a result of the open-loop controller design. Over time, minor disturbances accumulate, which cause the robot and rope of the counterweight to swing. After a human stabilizes the swinging motion, the robot can resume jumping for up to \SI{80}{\min}. 

\vspace{-10pt}
\section{DISCUSSION}
\label{sec_discussion}

\subsection{Space-qualification of components}

As mentioned in various sections, we designed \emph{SpaceHopper} towards operation in space by using appropriate materials (aluminum 7075 and carbon fiber), space-qualifiable motors, a differential drive design with crown gears, and lithium-ion batteries. However, the system is not fully space-graded yet. For example, the electronics are not radiation-shielded, and adequate thermal insulation is vital to survive in space. A first feasibility study regarding a thermal assessment for a lunar mission has already been done \cite{Trentini2023}. Furthermore, the system relies on an external motion capture system during testing. Therefore, an internal state estimation is needed that can estimate the robot's position and velocity relative to an inertial frame. The current sensor layout with three LRFs is inadequate to perform this task. Fusing an IMU (accelerometer and gyroscope), a camera, a LRF, and performing Range-Visual-Inertial Odometry (range-VIO) could be a possible solution \cite{xvio, Mars_Helicopter}.

\subsection{Safety and Robustness of DRL for Space Exploration}
The cost of space missions is significant, making safe and robust controllers a critical component of the system. Standard model-free RL has no theoretical safety guarantees. However, as seen in our experiments and prior work \cite{9981198}, DRL controllers, in practice, display excellent robustness even in highly complex environments. Thanks to modern training techniques such as domain randomization, resulting policies can generalize to unseen target environments. Furthermore, safe RL is an active field of research \cite{gu2023review}. In the future, such findings may be integrated into control approaches for additional theoretical guarantees.

\subsection{Simulating low-gravity scenarios}
\label{subsec:limitations_of_test_scenarios}
While the gimbal and the counterweight setup allow us to test the reorientation and jumping of SpaceHopper at Earth's gravity, they lack in accurately representing the robot's dynamics in low-gravity environments. The Earth's gravity still acts on the legs for both gravity offload systems. The inertia added through the gimbal test setup changes the system dynamics further. Additionally, imperfections in balancing the gimbal lead to constant torques acting on the robot, which increases the attitude error and leads to longer stabilization times. Testing only vertical jumping also falls short of validating pronking locomotion on real hardware. Moreover, the fast jumping dynamics cause slack in the counterweight rope, which introduces additional inaccuracies and makes sim to real transfer of DRL policies difficult. In future work, we plan a parabolic flight testing campaign that will overcome these limitations and further allow us to test the main locomotion goal and increase SpaceHopper's TRL.

\subsection{Adaptation to Irregular Terrain}
The irregular terrain of asteroids poses significant challenges for non-articulated hopping robots like MINERVA \cite{Yabuta2019-xs}. Unpredictable bouncing after landing severely limits precise locomotion. We hypothesize that a limbed system can adapt its legs for a soft landing, which allows for more controlled interactions with the surface. Future work will need to validate our control approach in this setting, especially on soft and granular media that may require different foot designs. 

\section{CONCLUSION AND FUTURE WORK}
\label{sec_conclusion}
In conclusion, \emph{SpaceHopper} is a research platform to investigate highly dynamic legged locomotion for exploring low-gravity celestial bodies, such as asteroids and moons. Unique features of the design, such as three legs, lightweight construction, small size, and a differential drive train, make the system well-adapted to jumping locomotion in low-gravity. In a zero gravity simulation, \emph{SpaceHopper} can change its attitude in \SI{1}{\s}. In a simulation of Ceres' gravity, it can also jump to commanded positions at a distance of \SI{6}{\m} with an average position error of \SI{0.316}{\m}. The real robot reaches an upright attitude within \SI{5}{\s} with a mean orientation error of \SI{9.7}{\deg} inside a gimbal test setup. The mechanical and electrical systems allow repeated vertical jumps in a counterweight setup. Due to the low-gravity optimized design of \emph{SpaceHopper}, creating a fitting test on Earth is a significant challenge. A natural follow-up of this work is to test \emph{SpaceHoppers'} capabilities in the microgravity environment of a parabolic flight. Further work is also necessary to show jumping on granular media, as often found on asteroids and moons. Finally, the state estimation must be extended and validated to work in the target environment.

\bibliographystyle{IEEEtran}
\balance
\bibliography{bibtex}

\end{document}